\documentclass[10pt,twocolumn,letterpaper]{article}

\usepackage[pagenumbers]{iccv} 

\definecolor{trf}{rgb}{0.5, 0.5, 1}
\definecolor{livingCoral}{HTML}{F36F63}

\usepackage{ulem}
\usepackage{multirow}
\usepackage{makecell}
\usepackage[thicklines]{cancel}
\usepackage{tcolorbox}
\usepackage{colortbl}
\usepackage{multicol}
\usepackage{bbding}
\usepackage{amsmath}

\definecolor{iccvblue}{rgb}{0.21,0.49,0.74}
\usepackage[pagebackref,breaklinks,colorlinks,allcolors=iccvblue]{hyperref}
\usepackage[accsupp]{axessibility}  

\def\paperID{10732} 
\def\confName{ICCV}
\def\confYear{2025}

\title{AG$^2$aussian: Anchor-Graph Structured Gaussian Splatting for Instance-Level 3D Scene Understanding and Editing}

\author{
Zhaonan Wang \quad Manyi Li\textsuperscript{†} \quad Changhe Tu\\
Shandong University\\
{\tt\small dyllanelliia\_wzn@163.com} \quad
{\tt\small manyili@sdu.edu.cn} \quad
{\tt\small chtu@sdu.edu.cn}
}

\begin{document}
\twocolumn[{%
 \renewcommand\twocolumn[1][]{#1}%
 \maketitle
 \begin{center}
   \centering
   \includegraphics[width=\textwidth]{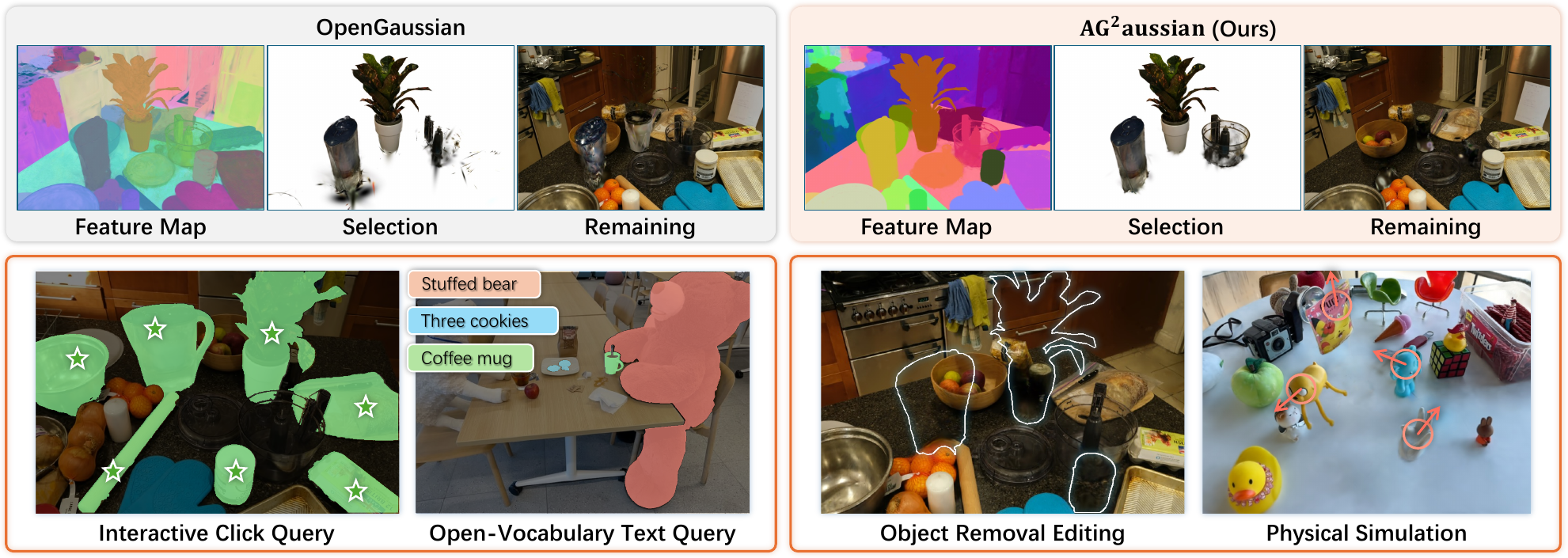}
   \captionof{figure}{We propose AG$^2$aussian, an anchor-graph structured Gaussian splatting for instance-level 3D scene understanding and editing tasks. Compared to existing works that attach semantic features to a collection of free Gaussians, we construct an anchor-graph structure to organize the semantic features and regulate the associated Gaussians, 
   resulting in a smooth feature distribution and a clean and accurate instance-level Gaussian selection (top row). Our approach benefits a series of applications, including the interactive click query, open-vocabulary text query, object removal editing, and physical simulation (bottom row).
   }
   \label{fig:teaser}
 \end{center}
 }]
 
\begingroup
  \renewcommand\thefootnote{}
  \footnotetext{\textsuperscript{†}Corresponding author.}
  \addtocounter{footnote}{-1}
\endgroup

\begin{abstract}
3D Gaussian Splatting (3DGS) has witnessed exponential adoption across diverse applications, driving a critical need for semantic-aware 3D Gaussian representations to enable scene understanding and editing tasks. Existing approaches typically attach semantic features to a collection of free Gaussians and distill the features via differentiable rendering, leading to noisy segmentation and a messy selection of Gaussians. In this paper, we introduce AG$^2$aussian, a novel framework that leverages an anchor-graph structure to organize semantic features and regulate Gaussian primitives. Our anchor-graph structure not only promotes compact and instance-aware Gaussian distributions, but also facilitates graph-based propagation, achieving a clean and accurate instance-level Gaussian selection. Extensive validation across four applications, i.e. interactive click-based query, open-vocabulary text-driven query, object removal editing, and physics simulation, demonstrates the advantages of our approach and its benefits to various applications. The experiments and ablation studies further evaluate the effectiveness of the key designs of our approach.
\end{abstract}    
\vspace{-15pt}
\section{Introduction}
\label{sec:intro}
3D Gaussian Splatting (3DGS)~\cite{kerbl3Dgaussians} models scenes as a collection of 3D Gaussian primitives and exhibits great utility in many applications~\cite{huang20242dgs,chen2024pgsr,chen2024gaussianeditor,xie2023physgaussian,zou2024triplane,zeng2024stag4d,li2024discene,conceptgraphs,zhang2024gaussiancube,huang2024dreamphysics}.
Recently, the advancements in large Vision-Language Models (VLMs)~\cite{radford2021learning,yang2024llm,cheng2023tracking,caron2021emerging,liu2024grounding,APE} have spurred a paradigm shift toward open-vocabulary scene understanding, where 3DGS serves as a bridge between high-fidelity scene representation and language-guided semantic reasoning. By combining these capabilities, researchers investigate open-vocabulary scene understanding with 3DGS~\cite{shi2024legaussian,qin2023langsplat,gaussian_grouping}, which further stimulates a wider range of semantic scene editing and manipulation tasks~\cite{gaussian_grouping,wu2024opengaussian,chen2024gaussianeditor,jiang2023gaussianshader}.

Existing GS-based scene understanding works~\cite{shi2024legaussian,zhou2024feature3dgs,qin2023langsplat,gaussian_grouping,wu2024opengaussian} distill semantic features from large VLMs~\cite{kirillov2023segment,radford2021learning,caron2021emerging,liu2024grounding} to free Gaussians via differentiable rendering, and use feature similarity to query the object-related Gaussians for visualization or editing purposes. However, these approaches face the challenge to obtain an accurate and clean Gaussian selection. As shown in Figure~\ref{fig:teaser}, when querying the objects, the selection often contains extra surrounding Gaussians and leaves the inner gaussian in the remaining scene, which apparently hinders the subsequent applications such as 3D scene editing and physical simulation.

The inaccurate and unclean Gaussian selection is due to the representation and rendering of 3DGS~\cite{kerbl3Dgaussians}. First, these approaches optimize unbounded Gaussians which tend to expand excessively across different views. It inadvertently causes semantic artifacts from unrelated objects and creates redundant features in localized regions. Second, the $\alpha$-blending rasterization back-propagates visual attributes along view rays but suffers from ambiguities when assigning the semantic features to overlapping gaussians, leading to inconsistent local features. Third, selecting Gaussians solely based on feature similarity ignores critical 3D spatial constraints, resulting in the inclusion of extra Gaussians from non-adjacent objects that share similar features. 

To address the above issues, we propose AG$^2$aussian, an anchor-graph structured Gaussian splatting for instance-level 3D scene understanding and editing tasks. The key idea is to construct an anchor-graph structure to organize the semantic features and regulate Gaussian primitives. Specifically, the anchor-graph acts as a higher-level semantic structure,
where each anchor is attached with a semantic feature and a small set of Gaussians. The advantages of this structure are two-fold: First, leveraging the semantic anchors to constrain the Gaussians, it promotes a compact and instance-aware Gaussian distribution. Second, the anchor-graph enables graph-based propagation to refine the semantic features of anchors, which significantly improves the accuracy and cleanness of object-related Gaussian selection.

Our technical contributions are summarized as follows: (1) We introduce \emph{the anchor-graph structured 3D Gaussian representation}, which constructs the anchor-graph structure to organize the semantic features and regulate Gaussian primitives for instance-level tasks. (2) We propose \emph{the anchor-graph feature propagation algorithm} to refine the semantic features and produce accurate Gaussian selections. (3) We employ the anchor-graph structured 3DGS to a series of \emph{scene understanding and editing applications}, which demonstrate the benefits of our representation for the processing operations of these applications. We conduct extensive experiments to validate the effectiveness of our approach, demonstrating superior performance in terms of Gaussian selection accuracy and the benefits to the intended applications.
Code released in \href{https://github.com/DyllanElliia/AGGaussian}{GitHub/AGGaussian}.



\section{Related Works}
\label{sec:related_work}

Numerous studies have adapted 3D Gaussian Splatting (3DGS)~\cite{kerbl3Dgaussians} for diverse applications~\cite{huang20242dgs,chen2024pgsr,dai2024high,guedon2024sugar,wu2024gaussianhead,chen2024gaussianeditor,gao2024towards,xu2024texturegs,yao2025refGS,xie2023physgaussian,waczynska2024games,mai2024everexactvolumetricellipsoid,wu2024recent}. For instance, 2DGS~\cite{huang20242dgs} projects 3D volumes into 2D oriented Gaussian disks to maintain view-consistent geometry reconstruction; Texture-GS~\cite{xu2024texturegs} and RefGaussian~\cite{yao2025refGS} disentangle geometry, texture, and reflections to enable flexible scene editing; EVER~\cite{mai2024everexactvolumetricellipsoid} introduces a physically accurate ellipsoid-based representation for physics-driven photorealistic rendering; PhyGaussian~\cite{xie2023physgaussian} incorporates physical properties into Gaussian primitives for high-fidelity simulation, to name a few.

On the other hand, the free-form nature of these Gaussians often leads to redundant components and loose spatial associations. Some works develop different data structures to organize the Gaussians to balance the efficiency and high-fidelity rendering~\cite{scaffoldgs,hac2024,fan2024lightgaussian,ren2024octree,gao2024mesh,wan2024superpoint}. Scaffold-GS~\cite{scaffoldgs} compresses the Gaussian representation using neural Gaussians to address the memory efficiency issue. Octree-GS~\cite{ren2024octree} utilizes the octree structure to enable adaptive detail control for complex large-scale scenes.

In addition to the geometry and appearance properties, some recent works study to enhance the 3D representations with semantic features~\cite{shi2024legaussian,zhou2024feature3dgs,qin2023langsplat,gaussian_grouping,cen2023saga,qu2024goi,wu2024opengaussian}. Leveraging the 3DGS representation, early works such as LEGaussian~\cite{shi2024legaussian} and Feature3DGS~\cite{zhou2024feature3dgs} laid the groundwork by integrating semantic information into the Gaussian framework to support semantic segmentation and open-vocabulary query. Subsequent works, including GaussianGrouping~\cite{gaussian_grouping}, SAGA~\cite{cen2023saga}, and OpenGaussian~\cite{wu2024opengaussian}, further advance this field by decoupling semantic cues from segmentation tasks. These methods enable instance-level segmentation and open-vocabulary object selection, but still face challenges: redundant Gaussian components, inconsistent feature assignments within local regions, and ambiguous object boundaries due to insufficient spatial constraints.

To address the remaining challenges, we construct the anchor-graph structure to organize the semantic features and regulate the Gaussian primitives. The closest works to ours are Scaffold-GS~\cite{scaffoldgs} and OpenGaussian~\cite{wu2024opengaussian}. Compared to them, our novel design lies in constructing the anchor-graph structure to regulate the explicit Gaussian primitives, and adopt the graph-based propagation to refine the semantic features. We demonstrate that these designs play a critical role in producing a clean and accurate Gaussian selection for a series of instance-level tasks.


\begin{figure*}
  \centering
  \includegraphics[width=\linewidth]{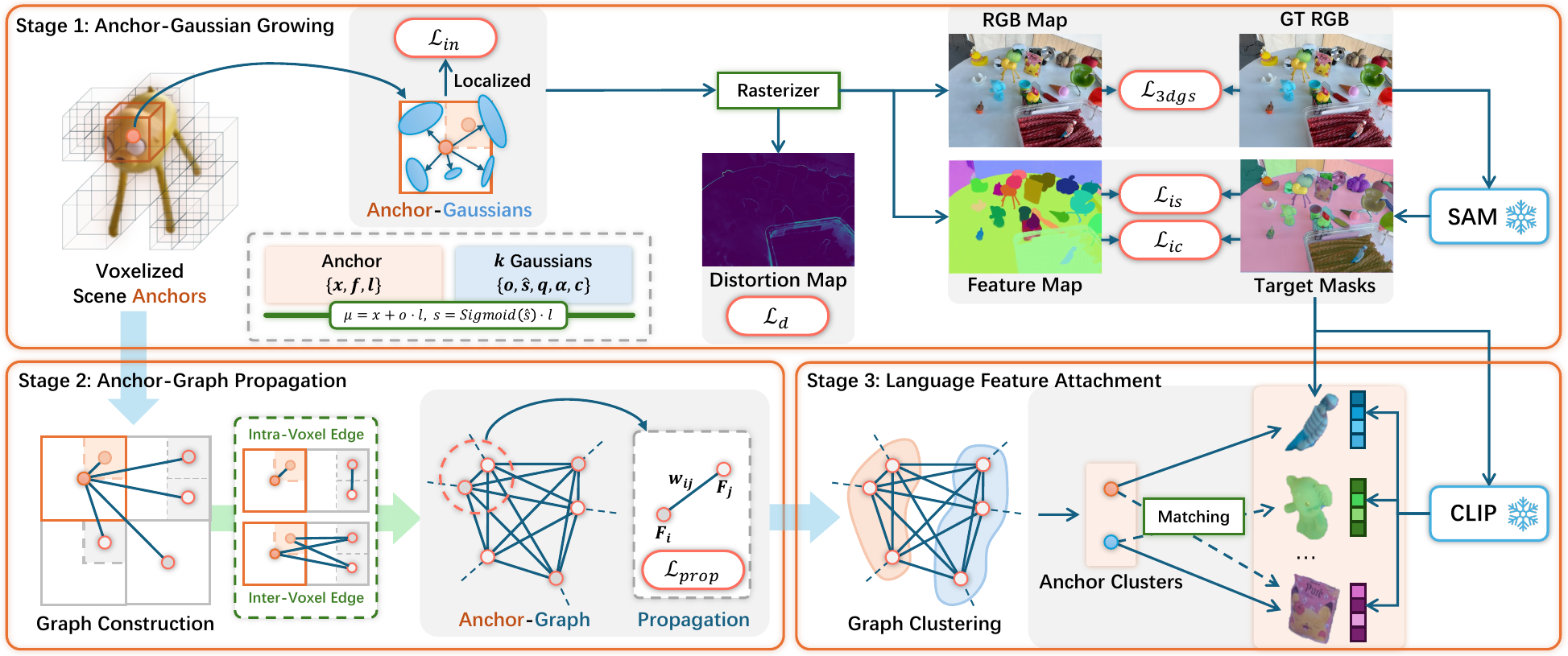}
  \caption{Our AG$^2$aussian consists of three stages. The first stage performs anchor-gaussian growing, where we initialize the anchors based on multi-resolution voxelization and optimize the anchor-Gaussian via a differentiable rasterizer. The second stage constructs the anchor-graph structure with sparse inter-voxel edges and intra-voxel edges, and adopts a graph-based propagation to refine the semantic features. Finally, the third stage localizes the object instances via graph clustering and attaches the matched language features to our representation.}
  \label{fig:pipeline}
\end{figure*}

\section{Method}

Given the multi-view images of a 3D scene, we optimize the anchor-graph structured Gaussian representation so that the renderings match the visual images and their semantic maps produced by large VLMs~\cite{kirillov2023segment,radford2021learning}. The entire pipeline is divided into three stages, as shown in Figure~\ref{fig:pipeline}. The first stage (Sec.~\ref{sec:ag_structure}) performs anchor-Gaussian growing to initialize the semantic anchors and the associated Gaussians. The second stage (Sec.\ref{sec:graph_propagation}) constructs the anchor-graph structure and adopts a graph-based propagation to refine the semantic features. The third stage (Sec.~\ref{sec:language features}) localizes the object instances and attaches language features to our anchor-graph structured Gaussian representation.

\subsection{Anchor-Gaussian Growing}
\label{sec:ag_structure}

The vanilla 3D Gaussian splatting~\cite{kerbl3Dgaussians} represents the 3D scenes with a collection of Gaussians $\mathcal{G}=\{g_i\}^{\mathcal{N}}_{i=1}$ where $\mathcal{N}$ is the Gaussian number.
Each Gaussian can be expressed as $g_i=\{\mu_i,s_i,q_i,\alpha_i,c_i\}$, with mean position $\mu_i\in\mathbb{R}^3$, scale factors $s_i\in\mathbb{R}^3$, quaternion $q_i\in\mathbb{R}^4$, opacity $\alpha_i\in\mathbb{R}$, and color properties $c_i\in\mathbb{R}^d$. The scaling and rotation form the covariance matrix of a Gaussian. These 3D Gaussians are splatted into a 2D image using differentiable rasterization via point-based $\alpha$-blending~\cite{zwicker2002ewa}. Given the camera viewpoint, the value at pixel $v$ of the rendered image is obtained by summing the Gaussians $\mathcal{N}_v$ intersected by its ray:
\begin{equation}
    I_v=\sum_{i\in{\mathcal{N}_v}}t_i c_i\prod^{i-1}_{j=1}(1-t_i),
\label{eq:gaussian}
\end{equation}
where $t_i = \alpha_i e^{-\frac 1 2(v-\hat{\mu}_i)\hat{\Sigma}_i^{-1}(v-\hat{\mu}_i)}$ is the contribution weight and $c_i$ is the color of the $i$-th Gaussian at pixel $v$. Moreover, by replacing the color values with the semantic feature $f_i$ attached to each Gaussian, we can obtain the feature map following the same rendering process.

However, the vanilla 3DGS~\cite{kerbl3Dgaussians} allows the Gaussians to freely extend and split within the 3D scene. It does not only cause inefficient data storage, but also affects the attribute back-propagation along the view rays, which becomes the main obstacle to obtaining a clean Gaussian selection. Inspired by works such as Scaffold-GS~\cite{scaffoldgs} and HAC~\cite{hac2024}, we develop a structured 3DGS that uses semantic anchors to regulate the corresponding Gaussians. The goal is to constrain the positions and scales of the Gaussians, in order to obtain a compact and instance-aware Gaussian distribution.

\noindent\textbf{Anchor-Gaussian Initialization.} Given the multi-view images, we reconstruct the sparse point cloud via Structure-from-Motion (SfM)~\cite{schonberger2016structure} as the initial positions of Gaussians. The sparse point cloud initializes a three-layer multi-resolution voxelization, where the centers of occupied voxels are taken as the locations of anchors $\mathcal{A}$.
Each anchor $a\in \mathcal{A}$ is parameterized by the center position $x\in\mathbb{R}^3$, a voxel size $l\in\mathbb{R}$, a semantic feature $f\in \mathbb{R}^3$, and $k$ child Gaussians $\{g_i=\{o_i,\hat{s}_i,q_i,\alpha_i,c_i\}\}^k_a$ each with an offset vector $o_i\in\mathbb{R}^3$ and relative scaling $\hat{s}_i$. Then the mean and scaling matrix of each Gaussian are:
\begin{equation}
    \mu_i=x+o_i\cdot l,\ s_i=Sigmoid(\hat{s}_i)\cdot l.
\label{eq:anchorgaussian}
\end{equation}
The Gaussians inherit the semantic feature of corresponding anchors, and can leverage the vanilla 3DGS rasterization to obtain the rendered images and feature maps.

\noindent\textbf{Anchor-Gaussian Densification.} The vanilla 3DGS densifies the Gaussians if they receive large gradients during training. Intuitively, as for our anchor-Gaussian structure, we densify the anchors and their associated Gaussians based on the gradients of Gaussians. Specifically, for any resolution layer, assuming a Gaussian $g$ is associated with the anchor $a$ located at the center of voxel $v$, if the Gaussian $g$ receives a gradient larger than a pre-defined threshold, we intend to create a new anchor and its associated Gaussians at the center of $v$. If there is no existing anchor positioned at this voxel center, we consider the new anchor initialization operation as valid. Otherwise, we cancel the initialization of this new anchor and continue. Note that the multi-layer voxelization enables a large resolution to place the anchors, thus ensuring the expressive ability of the anchor-Gaussian representation during training.

\noindent\textbf{Structured Spatial Regularization.} The anchor-Gaussian association enables the spatial regularization of Gaussians in terms of their positions and scaling. As demonstrated in Eq.~\ref{eq:anchorgaussian}, we restrict the scale of Gaussian to be within the voxel at which its corresponding anchor is placed. And we use a local constraint loss to regularize the positions:
\begin{equation}
    \mathcal{L}_{in}=\mathbb{E}_{o\in\{o_i\}^k_{a\in \mathcal{A}}}\parallel\exp({\rm ReLU}(\parallel o_i\parallel^2_2-1))\parallel
\end{equation}
where $\{o_i\}^k$ is the offset set of anchor's Gaussians.

We adopt the depth distortion loss from 2DGS~\cite{huang20242dgs} to suppress floating Gaussians and enhance the compactness of object surfaces by minimizing the distances between ray-Gaussian intersection points. The distortion at pixel $v$ is:
\begin{equation}
    \mathcal{L}_{d,v}=\sum_{i\in{\mathcal{N}_v}}\sum^{i-1}_{j=1}\omega_i\omega_j(z_i-z_j)^2
\end{equation}
where $\omega_i = t_i\prod^{i-1}_{j=1}(1-t_i)$ represents the contribution of the $i$-th Gaussian to the rendering results, and $z_i$ is the depth of this Gaussian along this ray.
The distortion loss is defined as $\mathcal{L}_d=\mathbb{E}\parallel \{\mathcal{L}_{d,v}\}^{H\times W} \parallel$.

\noindent\textbf{Semantic Contrastive Learning.}
\label{sec:contrastive_learning}
We use SAM~\cite{kirillov2023segment} to produce multi-view instance masks and distill the semantic feature for each anchor. The detailed masks that are entirely contained within other instances are removed.

We employ the contrastive learning strategy proposed by OpenGaussian~\cite{wu2024opengaussian} to encourage features rendered from the same object to be close to each other while those from different objects remain distant. For the $i$-th view of the multi-view images, given the rendered feature map $F$ and the binary mask $M_j\in\{0,1\}^{1\times H\times W}$ of the $j$-th object produced by SAM~\cite{kirillov2023segment}, we can obtain the mean feature within the mask region: $\overline{F}_j=(M_j\cdot F)/\sum M_j \in \mathbb{R}^3$. The intra-mask smoothing loss $\mathcal{L}_{is}$ and inter-mask contrastive loss $\mathcal{L}_{ic}$ are used to distill semantic features which are rendered to form the feature maps:
\begin{equation}
\begin{aligned}
\mathcal{L}_{is}&=\sum^{|M|}_{j=1}\sum^{H,W}_{h,w=1}M_{j,h,w}\parallel F_{:,h,w}-\overline{F}_j\parallel  \\
    \mathcal{L}_{ic}&=\frac{1}{m(m-1)}\sum^{|M|}_{j=1}\sum^{|M|}_{\substack{k=1 \\ k\ne j}}\frac{1}{\parallel \overline{F}_j-\overline{F}_k\parallel +1}
\end{aligned}
\end{equation}
where $|M|$ is the number of instance masks in current view, $\overline{F}_j$ and $\overline{F}_k$ denote the mean features of two different masks.

\subsection{Anchor-Graph Propagation}
\label{sec:graph_propagation}

After the first stage, we obtain the anchors and their associated Gaussians that visually exhibit good semantic features and appearance, but are far from a clean semantic representation for accurate object selections. There are two main reasons: First, due to occlusion between Gaussians, the features of their corresponding anchors within the same object can differ. Second, for objects that never co-appear in the same view, their features can still be similar. The above reasons make it difficult to distinguish the objects completely. 

We introduce the anchor-graph construction and propagation algorithm to address these issues. The main idea is to leverage local graph structures to refine the anchor features. We first construct the anchor graph based on their spatial distribution, then propagate to refine semantic similarities.

\noindent\textbf{Anchor Graph Construction.} Since the semantic anchors are placed at the centers of multi-resolution voxels, we connect the anchors based on the voxel neighborhood to construct the anchor graph. Specifically, for a voxel at the top layer, i.e. the layer with the smallest resolution, we collect all the anchors placed within this voxel, no matter which layer the anchors belong to. We create intra-voxel edges that connect any two of the anchors within a top-layer voxel, and inter-voxel edges between the anchors located within neighboring top-layer voxels.
In the following, we only allow feature propagation through these edges.

\noindent\textbf{Graph Laplacian Propagation.} 
We introduce a Gaussian-weighted graph Laplacian propagation to smooth the features of connected anchors while preserving sharp feature variations along object boundaries. Assuming a weighted adjacent matrix $W$ and the diagonal matrix ${\rm D}_{ii}=\sum_j {\rm W}_{ij}$, the Laplacian matrix is defined as
\begin{equation}
    {\rm L}={\rm D}-{\rm W},\quad {\rm L,D,W}\in \mathbb{R}^{|\mathcal{A}|\times|\mathcal{A}|}.
\end{equation}
Then the graph Laplacian propagation algorithm takes the Dirichlet Energy term as the propagation loss to enforce the smoothness between anchor features:
\begin{equation}\label{eq:L_prop}
    \mathcal{L}_{prop}=2{\rm\bf Tr}({\rm F^\top LF})=\sum_{i,j}w_{ij}\parallel {\rm F}_i-{\rm F}_j\parallel^2
\end{equation}
where ${\rm F}=\{f\}^{|\mathcal{A}|}\in\mathcal{R}^{|\mathcal{A}|\times 3}$ denotes the feature matrix of all anchors, and $F_i$ denoting the feature  of the $i$-th anchor. 

We set the weight $w_{ij}=\exp(-\frac{\parallel {\rm F}_i-{\rm F}_j\parallel^2}{2\tau^2})$ as a Gaussian kernel function related to the temperature $\tau=0.05$. It is worth noting that we don't need to explicitly construct the Laplacian matrix, but only the sparse non-zero $w_{ij}$ to compute the propagation loss.

\subsection{Language Feature Attachment}
\label{sec:language features}
After the second stage, we have already obtained an anchor-graph structured representation with semantic features to distinguish the objects. To further enable open-vocabulary query tasks, we perform graph-based clustering to localize the object instances and score matching to attach the additional language features to the anchors.

We use Union-Find~\cite{tarjan1975efficiency} for the graph-based clustering to efficiently identify and gather the connected anchors with similar features. Each cluster of anchors and the associate Gaussians form an object instance. By rendering the instance-related Gaussians with white color and the rest black color, we obtain the binary instance map $I_{\hat{a}}\in\mathbb{R}^{1\times H\times W}$, with $\hat{a}$ denoting the cluster of anchors.

The next step is to match the rendered instance maps and the ground-truth SAM masks for all the views, to obtain CLIP-encoded language features~\cite{radford2021learning} for the instances. Similar to \cite{wu2024opengaussian}, the matching score between $I_{\hat{a}}$ and $M_i$ is:
\begin{equation}
    S_{\hat{a},i}={\rm IoU}(I_{\hat{a}},M_i)\cdot(1-\parallel \overline{F}_{\hat{a}}-\overline{F}_i \parallel^1)
\end{equation}
where ${\rm IoU}$ is used to measure the overlap between $I_{\hat{a}}$ and $M_i$, while the second term estimates the distance between the mean features. Based on the scores, for the clustered anchors of each object, we select the most relevant mask by SAM~\cite{kirillov2023segment} and attach the feature $f^{clip}\in\mathbb{R}^{512}$ of corresponding CLIP~\cite{radford2021learning} encodings to these anchors.
Notably, to accelerate this process, we skip the clusters with only one single anchor, as they are mostly cluttered transparent points that do not contribute meaningfully to the scene.

\subsection{Training}
\label{sec:training}

Our three-stage training is set as follows:

\noindent\textbf{Stage 1: Anchor-Gaussian Growing.}
During the first 30k iterations, our approach focuses on the growing of anchors and the associated Gaussians. The loss for this stage is:
\begin{equation}
    \mathcal{L}_{stage1}=\mathcal{L}_{3dgs}+\lambda_{in}\mathcal{L}_{in}+\lambda_{is}\mathcal{L}_{is}+\lambda_{ic}\mathcal{L}_{ic}+\lambda_{d}\mathcal{L}_{d}
\end{equation}
with $\mathcal{L}_{3dgs}$ denotes the 3DGS reconstruction loss~\cite{kerbl3Dgaussians}.

\noindent\textbf{Stage 2: Anchor-Graph-based Feature Propagation.}
For the next 5k iterations, we first perform the anchor-graph construction, followed by the graph Laplacian propagation. The loss function is
\begin{equation}
\mathcal{L}_{stage2}=\mathcal{L}_{stage1}+\lambda_{prop}\mathcal{L}_{prop}
\end{equation}

\noindent\textbf{Stage 3: Language Feature Attachment.} As a post-processing stage, we sequentially perform the clustering, matching, and attach the mean language feature $f^{clip}$ of each object instance to the corresponding anchors. 

At the end of the training, we obtain the optimized anchor-graph structured 3D Gaussian splatting representation, where each anchor carries its position and corresponding voxel size $(x,l)$, the distilled semantic feature $f$, the language feature $f^{clip}$, and a small set of associated Gaussians $\{g_i\}^k$. Note that the anchor-graph structure not only makes a better scene understanding with the semantic features, but also ensures a compact set of Gaussians and alleviates the computation burden. 


\section{Instance-level Application}
\label{sec:interactive}
The clean semantic Gaussian representation produced by our approach benefits a series of 3D scene understanding and editing applications. Below we describe how to deploy our approach for various applications.

\noindent\textbf{Click Query.}
We adopt a local region-growing strategy for the click query application. Given the camera viewpoint of the image, we render the 3D scene into a depth map. The depth allows us to transform the clicked pixel into a 3D point $p$ with the camera's extrinsic and intrinsic matrices. Next, using a Nearest Neighbor search, we identify the closest anchor  $a^*$ as the initial seed anchor. From this seed, the region grows iteratively: if an anchor $a_i$ is connected to another anchor $a_j$ with an edge weight $w_{ij} > 0.90$, then $a_j$ should be included in the selected region. Note that we double the voxel size as the threshold for creating the graph edges only during this query. 

\begin{figure*}[!htbp]
  \centering
  \includegraphics[width=1.0\linewidth]{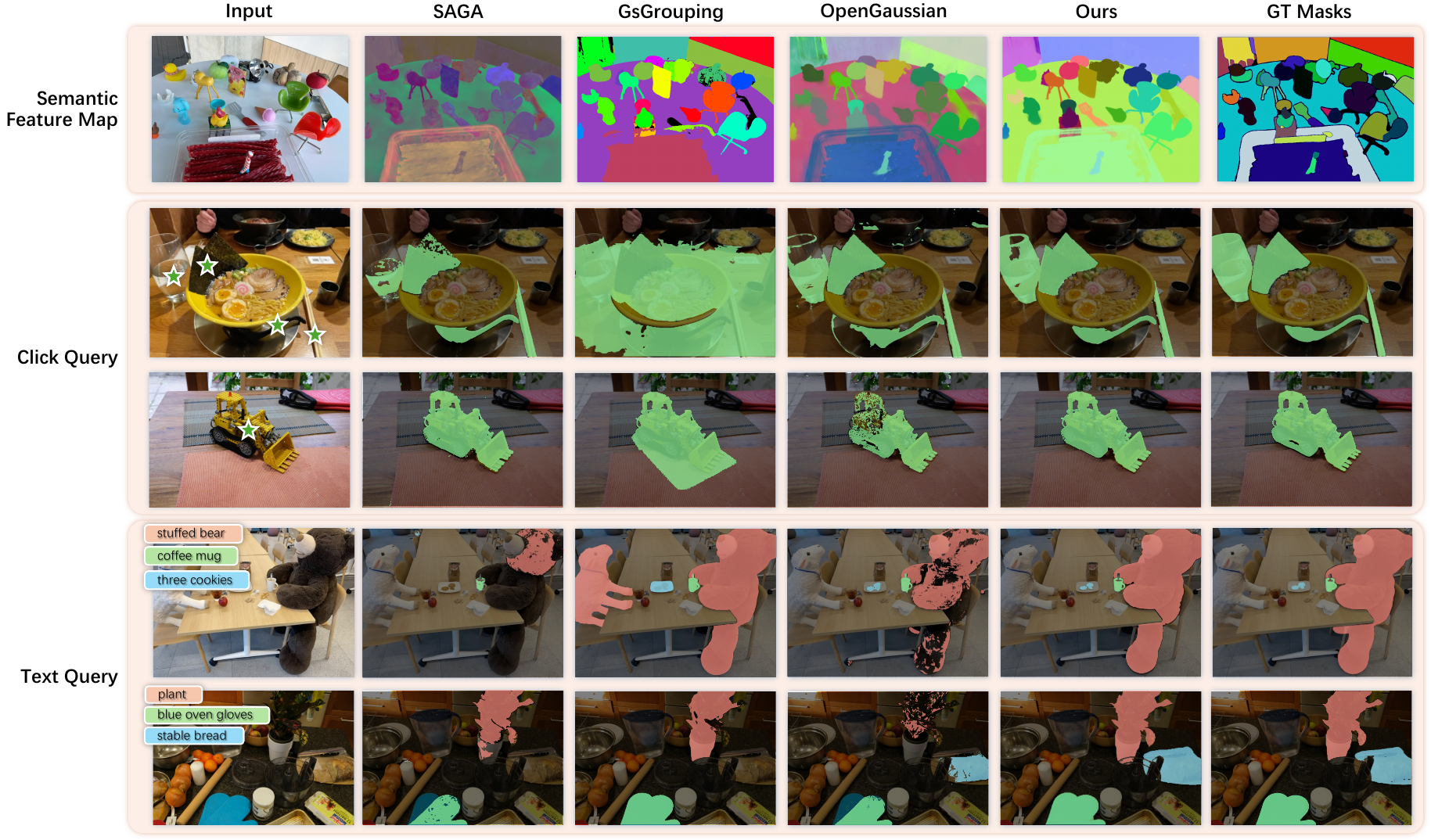}
  \caption{The rendered semantic feature map and object query results of our approach and related methods. The presented GT images are generated by SAM~\cite{kirillov2023segment}. By contrast, our approach produces more precise and less noisy instance masks for the queried objects.
  }
  \label{fig:exp_query}
\end{figure*}

\noindent\textbf{Text Query.}
The text query adopts the same growing strategy with click query but a slightly different seed initialization. Given an open-vocabulary text query, we extract its text embedding using CLIP encoder~\cite{radford2021learning} and calculate the cosine similarity between this feature and the language features of clustered anchors.
Assuming  $\varepsilon$ to denote the maximum feature similarity score among them, we select the anchor clustering whose language features have a similarity larger than $\varepsilon-0.1$ as the initial seed anchors. By performing the region-growing process, we obtain the anchors within the region as the query results. 

\noindent\textbf{Scene Editing.}
A clean instance-aware Gaussian representation enables object selection and local scene editing operations. Specifically, for 3D object removal, directly deleting the selected anchors and the associated Gaussians (with the above query techniques) would leave artifacts in the region. Therefore, we add a small number of new anchors within the object region and render the image with a region mask.  The region mask is obtained by rendering the Gaussians of the newly added anchors and their neighbor anchors. Then we perform 2D image inpainting with LaMa~\cite{suvorov2022resolution} to obtain realistic images and optimize our anchor-graph structured Gaussians w.r.t. the inpainted images.

\noindent\textbf{Physical Simulation.}
Physical simulation aims to imitate the physical interactions between the selected object and the remaining scene. We adopt PhyGaussian~\cite{xie2023physgaussian}, a Gaussian-based simulator implemented via MLS-MPM~\cite{hu2018mlsmpmcpic}, as our physical engine. The selected object is assigned Young’s modulus $E=2e^{8}$ and Poisson’s ratio $\nu=0.4$ to prevent deformation during simulation, while the remaining scene has lower physical coefficients ($E=2e^{6}$, $\nu=0.3$). More implementation details of the physical simulation application is presented are the supplementary material.
\section{Experiments}

\subsection{Setup}

\begin{table*}[!ht]
\centering
\caption{Quantitative evaluation of object query applications on LERF dataset~\cite{kerr2023lerf}.
}
\resizebox{\linewidth}{!}{
\begin{tabular}{c|c|ccccc|ccccc}
\toprule
& & \multicolumn{5}{c|}{mIoU $\uparrow$} 
& \multicolumn{5}{c}{mBIoU. $\uparrow$} \\
\multirow{-2}{*}{Query}&\multirow{-2}{*}{Methods}&
  \texttt{figurines} &
  \texttt{teatime} &
  \texttt{ramen} &
  \texttt{kitchen} &
  \cellcolor[HTML]{EFEFEF}\textbf{Mean} &
  \texttt{figurines} &
  \texttt{teatime} &
  \texttt{ramen} &
  \texttt{kitchen} &
  \cellcolor[HTML]{EFEFEF}\textbf{Mean} \\ 
  \midrule
\multirow{4}{*}{Click}
&SAGA~\cite{cen2023saga}            &81.16 &92.91 & 63.08 &\textbf{83.03} & \cellcolor[HTML]{EFEFEF}80.05 & 76.21 &60.27 &54.03 &  \textbf{62.07} & \cellcolor[HTML]{EFEFEF}63.15 \\
&GsGrouping~\cite{gaussian_grouping}      &32.54 &75.06 & 39.99 &14.70 & \cellcolor[HTML]{EFEFEF}40.57 & 28.43 &57.20 &35.86 &  9.74  & \cellcolor[HTML]{EFEFEF}32.81 \\
&OpenGaussian~\cite{wu2024opengaussian}    &85.15 &80.31 & 48.88 &79.48 & \cellcolor[HTML]{EFEFEF}73.46 & 81.28 &51.34 &50.73 &  49.95 & \cellcolor[HTML]{EFEFEF}58.33 \\
&Ours &
  \textbf{88.61} &
  \textbf{95.87} &
  \textbf{88.04} &
  {80.80} &
  \cellcolor[HTML]{EFEFEF}\textbf{88.33} &
  \textbf{85.27} &
  \textbf{87.65} &
  \textbf{71.54} &
  {49.86} &
  \cellcolor[HTML]{EFEFEF}\textbf{73.58} \\
\midrule
\multirow{4}{*}{Text}
&SAGA~\cite{cen2023saga}            & 16.76 & 18.90 & 11.01  & 5.29  & \cellcolor[HTML]{EFEFEF} 12.99 & 16.14 & 17.74 & 10.20 &   3.58 & \cellcolor[HTML]{EFEFEF} 11.92 \\
&GsGrouping~\cite{gaussian_grouping}      & 28.11 & 64.71 &  33.47 & 16.68 & \cellcolor[HTML]{EFEFEF} 35.74
                & 27.19 & 59.30 & 31.82 &  14.82 & \cellcolor[HTML]{EFEFEF} 33.28 \\
&OpenGaussian~\cite{wu2024opengaussian}    & 57.41 & 62.54 &  30.77 & 25.96 & \cellcolor[HTML]{EFEFEF} 44.17 
                & 54.83 & 56.38 & 26.94 & 17.32 & \cellcolor[HTML]{EFEFEF} 38.87 \\
&Ours &
  \textbf{66.98} &
  \textbf{71.62} &
  \textbf{47.99} &
  \textbf{30.82} &
  \cellcolor[HTML]{EFEFEF}\textbf{54.35} &
  \textbf{65.30} &
  \textbf{67.83} &
  \textbf{42.45} &
  \textbf{22.15} &
  \cellcolor[HTML]{EFEFEF}\textbf{49.43} \\
  \bottomrule
\end{tabular}
}
\label{tab:lerf}
\end{table*}

\noindent\textbf{Datasets.}
We evaluate our approach on the LERF-OVS~\cite{kerr2023lerf} datasets for the open-vocabulary object query, scene editing, and physical simulation applications. The four scenes are manually annotated with SAM to obtain accurate masks instead of the coarse bounding boxes.
We further evaluate object querying on Mip-NeRF360~\cite{barron2022mip} and object selection on LLFF~\cite{mildenhall2019llff}, which contain more challenging examples.

\noindent\textbf{Implementation Details.}
We use the SAM ViT-H model to generate 2D masks of input images, and extract language features for each instance using the OpenCLIP ViT-B/32 model.
Throughout all experiments, each anchor is assigned $k=5$ Gaussians.
For multi-resolution voxels, we develop three resolution levels, each with a voxel size 4 times larger than the previous one. The minimum resolution (for the top layer) is $200^3$.
During training, we use Adam optimizer and set $\lambda_{in}=0.5$, $\lambda_{is}=2.5$, $\lambda_{ic}=0.25$, $\lambda_{d}=50$, $\lambda_{prop}=0.01$.
For the scene editing task, we use LaMa Big-Lamm to generate 2D inpainting images.

\subsection{Scene Understanding and Object Query}
\begin{figure}[!htbp]
  \centering
  \includegraphics[width=\linewidth]{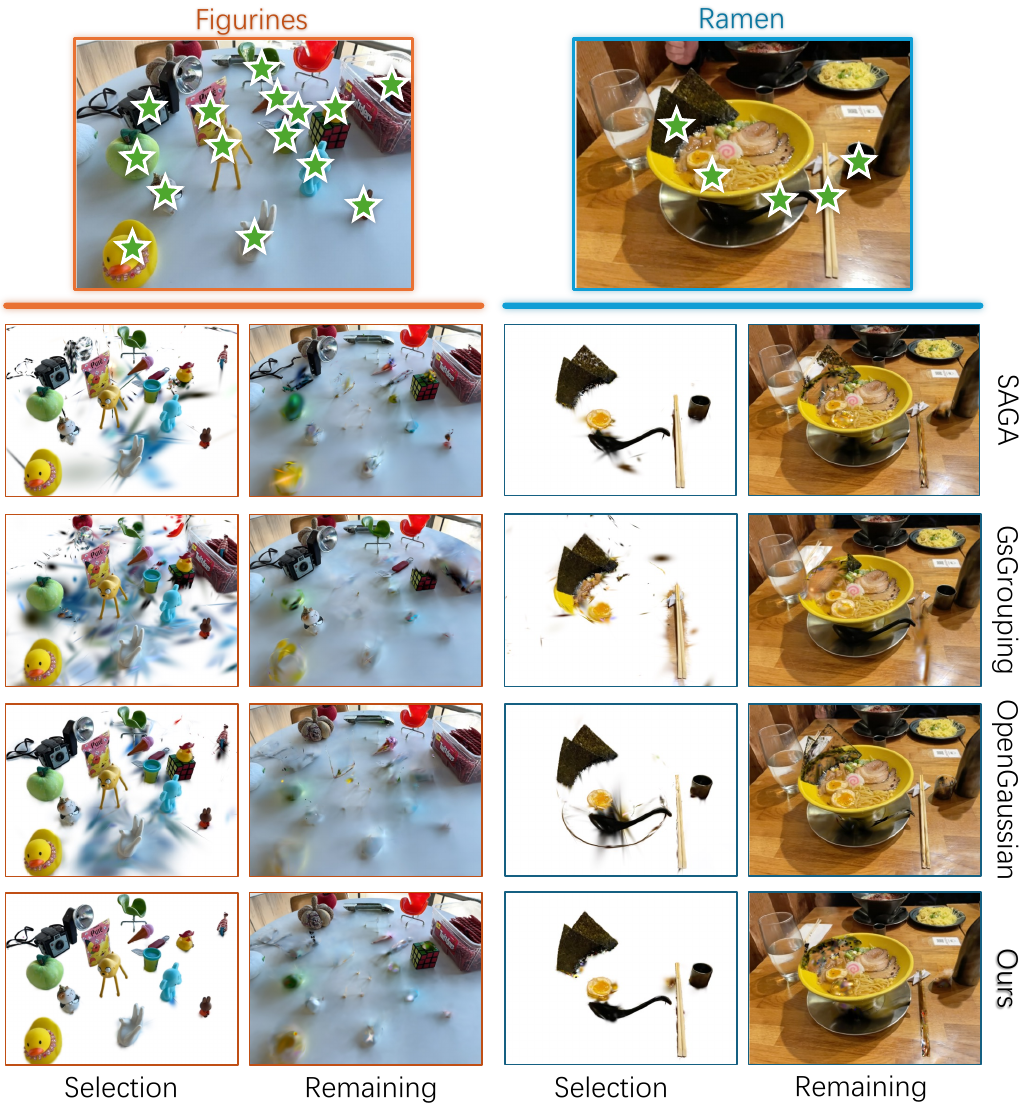}
  \caption{Click-based instance-level Gaussian selection and the remaining scenes. Our approach makes more accurate and complete Gaussian selections, though with visual artifacts of occluded Gaussians in the remaining scenes.
  }
  \label{fig:exp_click}
\end{figure}

We conduct the scene understanding and open-vocabulary object query tasks and compare to the existing state-of-the-art 3DGS-based approaches, i.e. SAGA~\cite{cen2023saga}, GsGrouping~\cite{gaussian_grouping}, OpenGaussian~\cite{wu2024opengaussian}. The object query tasks include click query and text query with object class names.

Figure~\ref{fig:exp_query} shows the rendered feature maps and the results of click query and text query for a variety of scenes. As shown in Figure~\ref{fig:exp_query}, our approach produces smooth and clean semantic feature maps that can easily distinguish the objects with accurate boundaries. This leads to more precise query results, compared to the noisy masks and blurry boundaries with other approaches. Please refer to the supplementary material for more results.

Table~\ref{tab:lerf} reports the quantitative evaluation. We use average IoU and BoundaryIoU between the binary images rendered with the selected anchor-Gaussians and the ground-truth object mask as the evaluation metrics. Our results achieve remarkable improvement over the others in terms of both the click query and text query, validating the high accuracy of the mask region and object boundaries.







\subsection{3D Scene Editing and Physical Simulation}

An accurate selection of the specified object is the prerequisite for the scene manipulation tasks. Unlike object query tasks that only require precise object masks for 2D visualization, object selection emphasizes accurately picking the Gaussians for specific object, including the inner Gaussians.

\begin{figure}[!t]
  \centering
  \includegraphics[width=\linewidth]{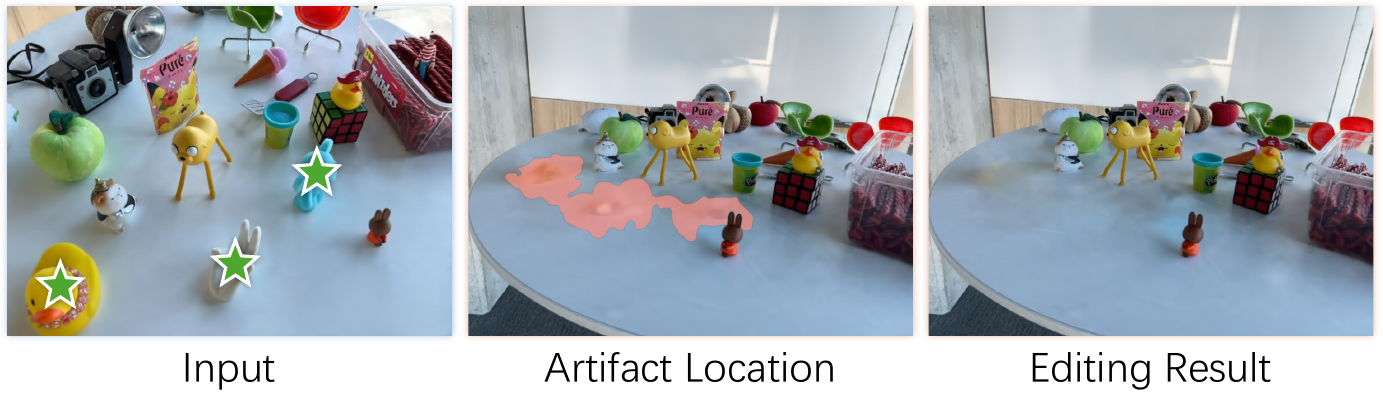}
  \caption{Object removal results. After deleting the selected Gaussians, we localize artifact regions and inpaint the remaining scene.
  }
  \label{fig:exp_inpainting}
\end{figure}
\textbf{}
\begin{figure*}[!ht]
  \centering
  \includegraphics[width=0.9\linewidth]{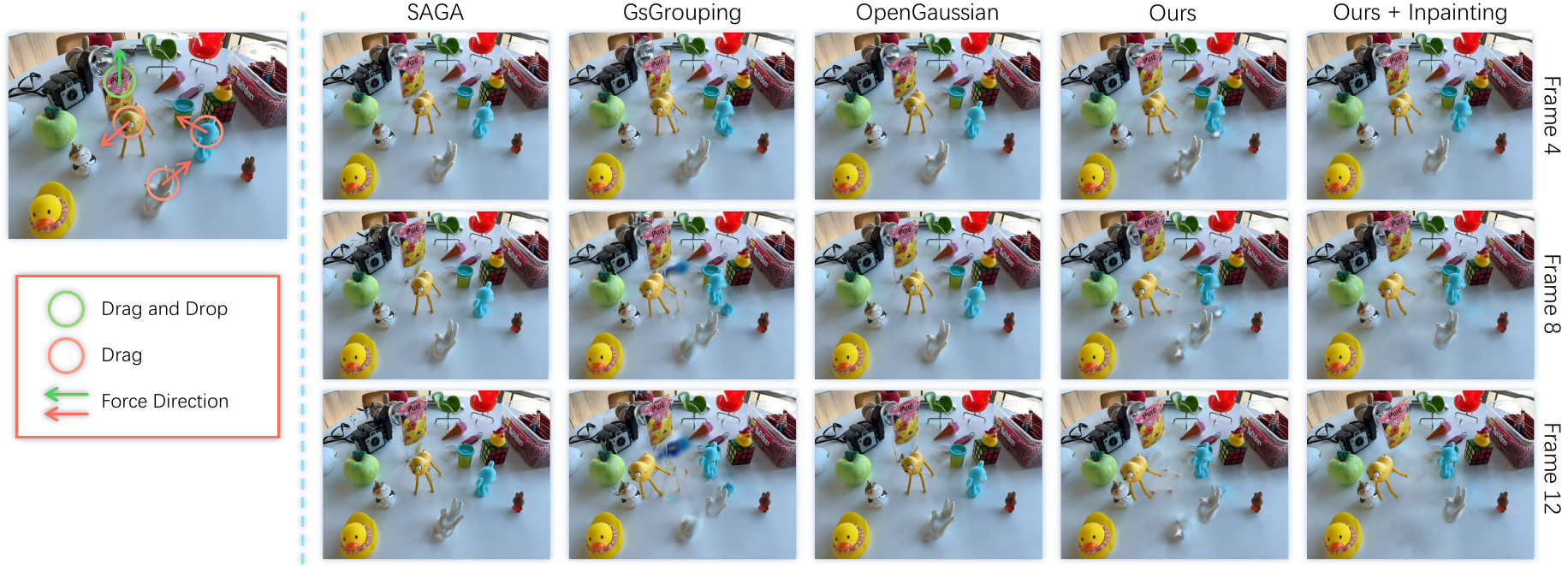}
  \caption{Physical simulation by applying the external forces (red arrows) to drag the objects. All methods are simulated with identical physical coefficients using PhysGaussian. The clean and accurate Gaussian selection enables the realistic simulation with our approach.
  }
  \label{fig:exp_simulation}
\end{figure*}

\noindent\textbf{Object Selection.}
We present the object selection results of ours and other existing methods in Figure~\ref{fig:exp_click}. Their default parameters and operation processes are adopted in the experiments. As for our approach, we perform the click query and take the Gaussians associated with the selected anchors.

In Figure~\ref{fig:exp_click}, we render both the selected and the remaining Gaussians. In spite of the clean selection shown in the rendering of the selected Gaussians, we can see that the objects are more completely removed in the remaining scenes, e.g. the white hand in the Figurines scene and chopsticks in the Ramen scene. Although there are still visual artifacts in the remaining scenes, it is because the remaining Gaussians of the other objects don't have the correct appearance due to the occlusion problem of multi-view reconstruction.

\noindent\textbf{3D Scene Editing.}
We further present the 3D object removal editing results to validate our claim that our approach exceeds an accurate and complete object selection. Since our anchor-graph structure can accurately localize the anchors corresponding to the object and its extended boundary, we can easily identify the artifact region to be repaired without affecting the surrounding objects.

\noindent\textbf{Physical Simulation.}
The physical simulation task poses a strict requirement for the object selection, since the object Gaussians entangled with the remaining scene would cause unrealistic simulation results. Figure~\ref{fig:exp_simulation} shows the simulation results of ours and other approaches. Taking the upward-dragged bag as an example, it remains in the original position in the SAGA~\cite{cen2023saga} and OpenGaussian~\cite{wu2024opengaussian} results, because most of the inner-object Gaussians are not selected. As for GaussianGrouping~\cite{gaussian_grouping}, which uses a convex hull geometry to select the Gaussians, it moves the bag but leaves messy surrounding Gaussians. By contrast, our approach not only successfully drags the bag, but also leaves small appearance artifacts that be easily localized and inpainted with our anchor-graph structure.

\subsection{Ablation Study.}
\begin{figure}[!t]
  \centering
  \includegraphics[width=\linewidth]{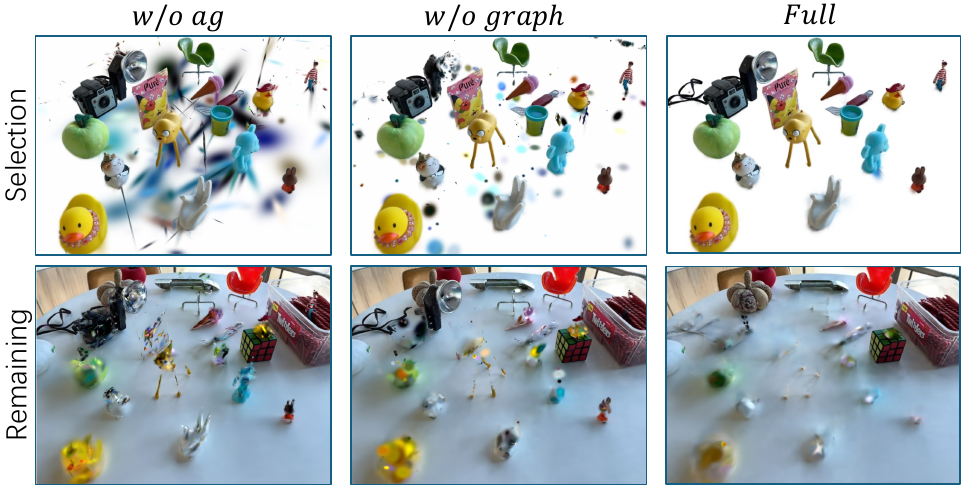}
  \caption{Visual results of the ablation study for anchor-graph structure design. With the anchor-graph structure and the graph-related operations, our full approach produces precise Gaussian selections and a relatively clean remaining scene.
  }
  \label{fig:exp_ab_small}
\end{figure}

\noindent\textbf{Anchor-Graph Structure Design.}
We emphasize the role of the anchor-graph structure in organizing the Gaussians for a clean object selection. The simplest baseline $w/o\ ag$ is to remove the anchor-structure from our approach. That is, the semantic features are attached to the vanilla Gaussians~\cite{kerbl3Dgaussians} and learned from contrastive learning. Another baseline $w/o\ graph$ is to retain the anchor-gaussian structure but omit all the graph-related operations, i.e. the graph propagation in the second stage and the region growing in the query application. The implementation of this baseline is to remove the loss $\mathcal{L}_{prop}$ and use the semantic feature similarity for the query applications. The qualitative results are shown in Figure~\ref{fig:exp_ab_small}.

\begin{table}[!t]
\centering
\caption{Quantitative evaluations of the ablation study for the two graph-related operations.
}
\resizebox{0.8\linewidth}{!}{
\begin{tabular}{c|cc|cc}
\toprule
Case & $\mathcal{L}_{prop}$     & $GraphSeg$        & mIoU $\uparrow$   & mBIoU. $\uparrow$  \\
\midrule
\#1  &                          &                   & 42.72             & 40.20             \\
\#2  &                          & \Checkmark        & 45.85             & 43.51             \\
\#3  & \Checkmark               &                   & 49.10             & 44.72             \\
Full & \Checkmark               & \Checkmark        & \textbf{54.35}             & \textbf{49.43}             \\
\bottomrule
\end{tabular}
}
\label{tab:ab_graph}
\end{table}
\noindent\textbf{Graph-Related Operations.} We further analyze the effects of each graph-related operation separately. The quantitative evaluation is reported in Table~\ref{tab:ab_graph}, where the two variables $\mathcal{L}_{prop}$ and $GraphSeg$ determine the involvement of the graph propagation in the second stage and the region growing in the query application, respectively. Table~\ref{tab:ab_graph} validates the significant effectiveness of the two graph-related operations in making accurate object queries.






\section{Conclusion}
We present AG$^2$aussian, an anchor-graph structured Gaussian representation for instance-level 3D scene understanding and editing tasks. The key idea is to construct an anchor-graph structure to organize the semantic features and regulate the associate Gaussians to produce a compact and instance-aware Gaussian distribution. The graph-related operations facilitate to make clean and accurate instance-level Gaussian selection, which, as demonstrated with our experiments, exhibit great benefits to the scene understanding and editing applications, including interactive click query, open-vocabulary text query, localized scene editing, and physical simulation.

Our approach still holds some limitations. Similar to the existing approaches, our method suffers from the glass and metal objects due to the feature blending. And sometimes the over-segmentation of SAM introduces additional difficulty to maintaining the consistent features of the objects.

\section{Acknowledgement}
This work is supported by the Joint Funds of the National Natural Science Foundation of China (U23A20312), the Excellent Young Scientists Fund Program (Overseas) of Shandong Province (No.2023HWYQ-034), the National Natural Science Foundation of China (62302269), and a grant from the Natural Science Foundation of Shandong Province (No.ZR2023QF077).
{
    \small
    \bibliographystyle{ieeenat_fullname}
    \bibliography{main}
}
\clearpage
\setcounter{page}{1}
\maketitlesupplementary
\appendix

In this supplementary material, we first present the implementation detail of the physical simulation task in Sec.~\ref{appendix:physical}.
Then, we present more object query comparisons in Sec.~\ref{appendix:text} and Sec.~\ref{appendix:click}. In Sec.~\ref{appendix:edit}, we demonstrate the robustness of our editing method and provide additional editing results on two scenes from the Mip-NeRF360 dataset~\cite{barron2022mip}. Finally, in Sec.~\ref{appendix:ab} and Sec.~\ref{appendix:structured}, we provide more ablation study results, including both qualitative and quantitative analysis.



\section{Application of Physical Simulation}\label{appendix:physical}

In our experiment, we adopt PhyGaussian~\cite{xie2023physgaussian}, a Gaussian-based simulator implemented via MLS-MPM~\cite{hu2018mlsmpmcpic}, as our physical engine. The Gaussians are regarded as particles to perform the simulation.
For computational efficiency purposes, we remove the background using a bounding box and retain only the foreground particles whose opacity $\alpha>0.02$ for simulation. 
Specifically, in our experiments, we first use a query operation to select the object to be simulated. This object is then assigned Young’s modulus $E=2e^{8}$ and Poisson’s ratio $\nu=0.4$ to prevent deformation during simulation.
The remaining particles within the bounding box, which serve as sticky boundary conditions with lower physical coefficients ($E=2e^{6}$, $\nu=0.3$), enable the simulated object to be easily separated from the surroundings. All of these particles are subsequently discretized into a grid $64^3$.
For all the physical simulation experiments, we simulate a total of 30 frames.
All particles in this application are assigned von Mises Plasticity material.

\begin{figure}[!ht]
  \centering
  \includegraphics[width=\linewidth]{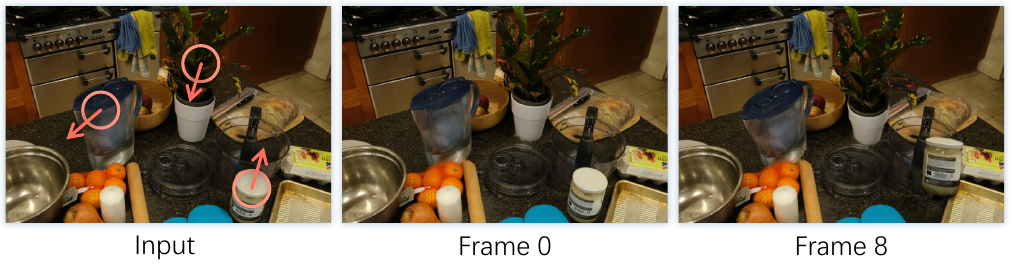}
  \caption{Physical simulation by applying the external forces (red arrows) to drag the selected objects.
  }
  \label{fig:appendix_simulation}
\end{figure}

\section{More Text Query Results}\label{appendix:text}

\begin{figure*}[ht]
  \centering
  \includegraphics[width=0.9\linewidth]{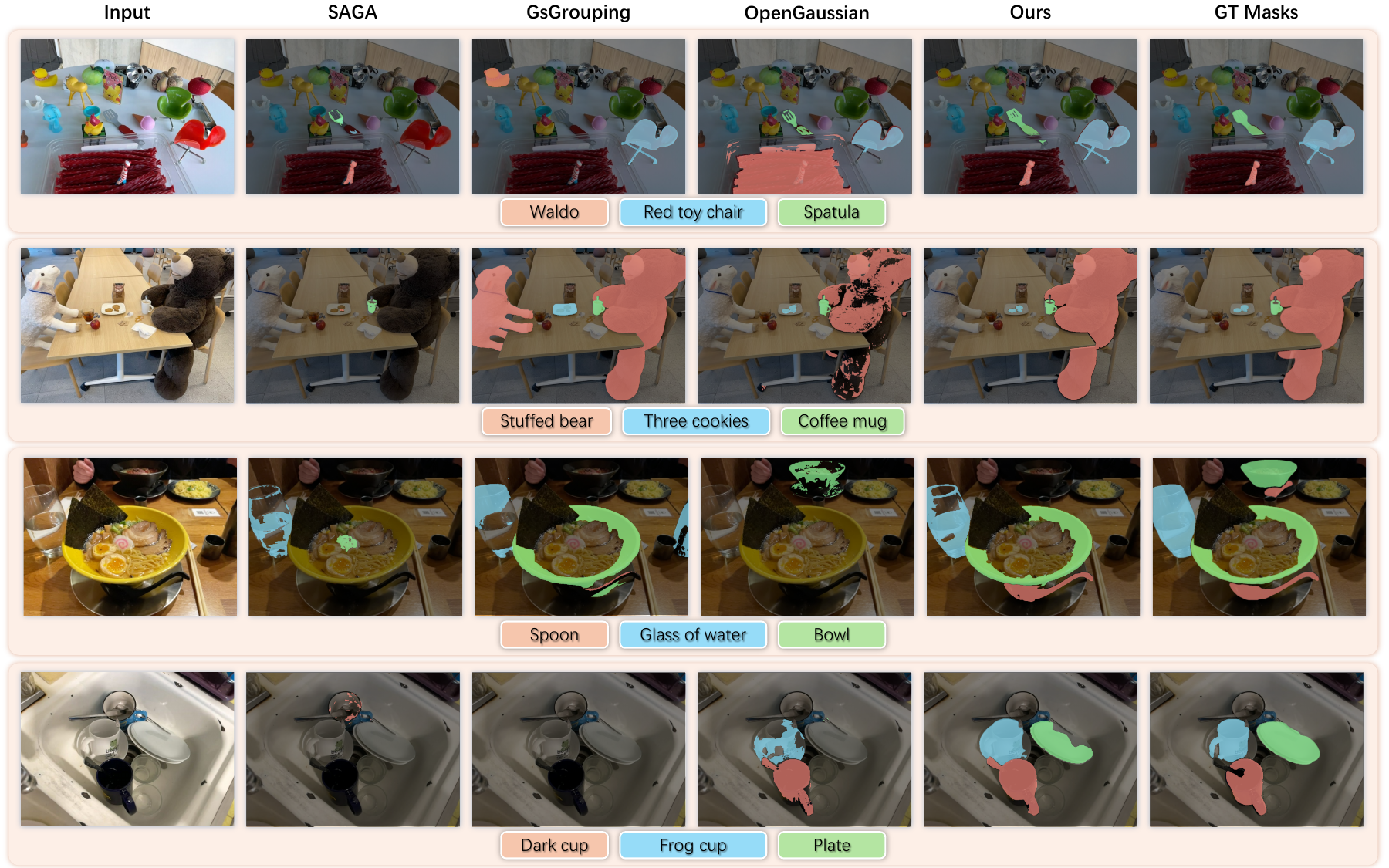}
  \caption{Open-vocabulary 3D object selection on the LERF dataset~\cite{kerr2023lerf}. AG$^2$aussian outperforms other approaches in accurately identifying the clean and complete 3D objects corresponding to text queries.}
  \label{fig:app_text}
\end{figure*}

We visualize more results of the open-vocabulary text query task in Figure~\ref{fig:app_text}, where our method demonstrates a clear advantage in selecting the complete 3D objects. By contrast, OpenGaussian~\cite{wu2024opengaussian}, due to its codebook-based clustering approach, often fails to group an entire object into a single cluster, as seen with the "waldo" in the first row and the "stuffed bear" in the second row. Similarly, GsGrouping~\cite{gaussian_grouping} frequently includes incorrect object IDs for the query, as seen with the "stuffed bear" in the second row and the "glass of water" in the third row. Meanwhile, SAGA~\cite{cen2023saga} uses a limited number of clusters and is less aware of spatial information, making it prone to missing matches and selecting incorrect regions.

\begin{figure*}[ht]
  \centering
  \includegraphics[width=0.9\linewidth]{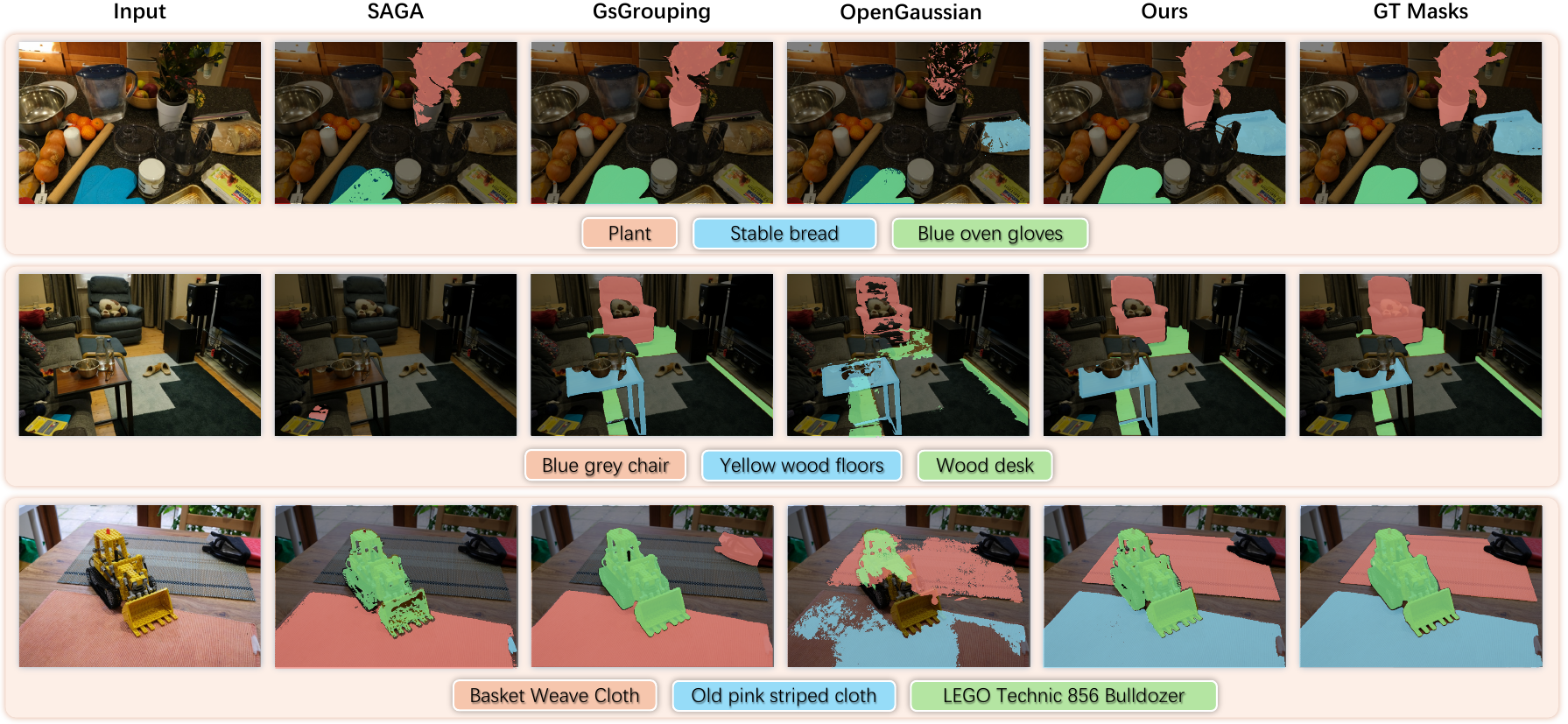}
  \caption{Open-vocabulary 3D object selection on the Mip-NeRF360 dataset~\cite{barron2022mip}.}
  \label{fig:app_miou_text}
\end{figure*}

In Table~\ref{tab:text_mip360} and Figure~\ref{fig:app_miou_text}, we further report both the quantitative and qualitative results of open-vocabulary querying on Mip-NeRF360~\cite{barron2022mip}, evaluated with the vocabulary provided by LEGaussian~\cite{shi2024legaussian}. Our results consistently outperform existing approaches, achieving significant improvements in both mIoU and mBIoU. These gains hold across diverse scenes and object types, and are especially observed on thin, partially occluded, or clutter-surrounded objects. Qualitative results further validate that our selected regions can align well with the entire instance, whereas others always leave fragmented or jagged boundaries.

\section{More Click Query Results}\label{appendix:click}

\begin{figure*}[ht]
  \centering
  \includegraphics[width=0.9\linewidth]{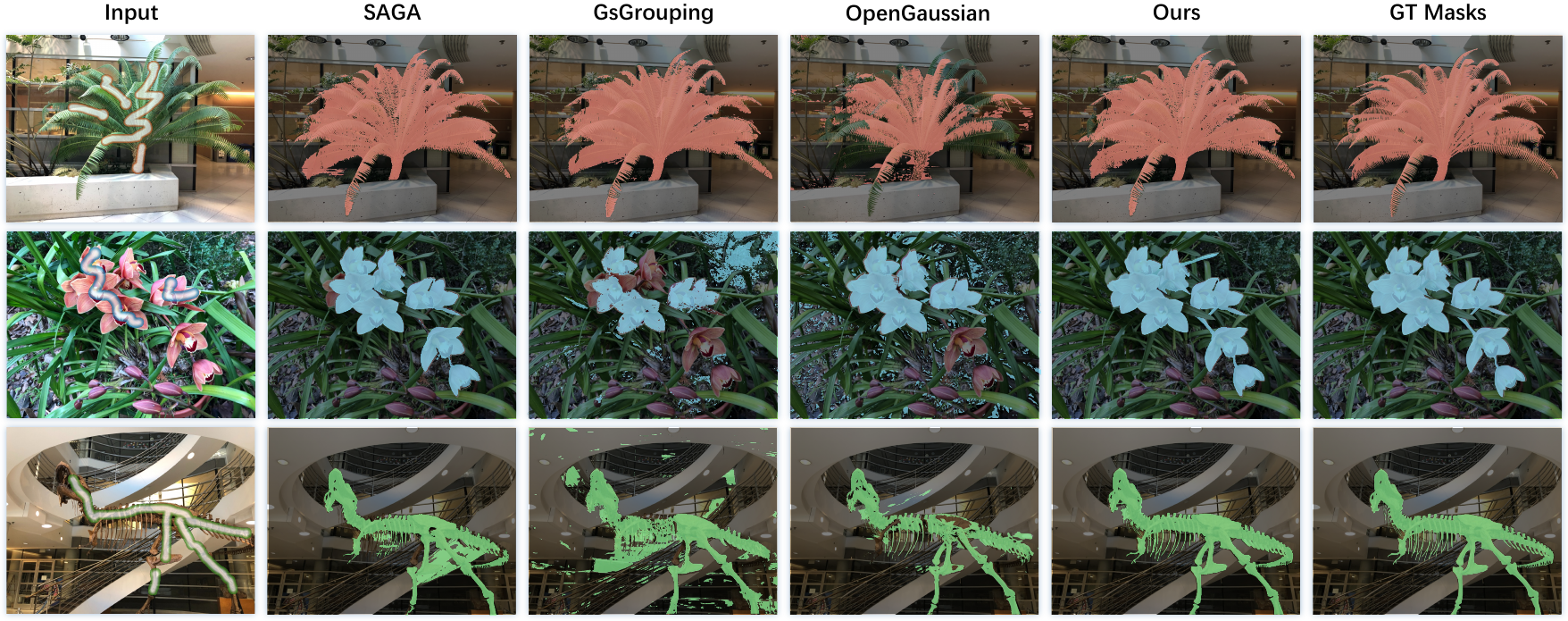}
  \caption{Scribbles-based 3D object selection on the LLFF dataset~\cite{barron2022mip}.}
  \label{fig:app_click_llff}
\end{figure*}

We report more object selection results on LLFF~\cite{mildenhall2019llff} in Figure~\ref{fig:app_click_llff} and Table~\ref{tab:click_llff}, using the scribbles provided by NVOS~\cite{ren-cvpr2022-nvos}.
As input, we first shrink the scribbles into skeleton lines and then use the pixels on the skeleton as click query points.
By contrast, our method yields more accurate segmentation for complex objects like fern and dinosaur fossils, benefiting from the use of localized anchor-Gaussian and our anchor-graph-based strategy.

\section{More Object Editing Results}\label{appendix:edit}

\begin{figure}[!t]
  \centering
  \includegraphics[width=\linewidth]{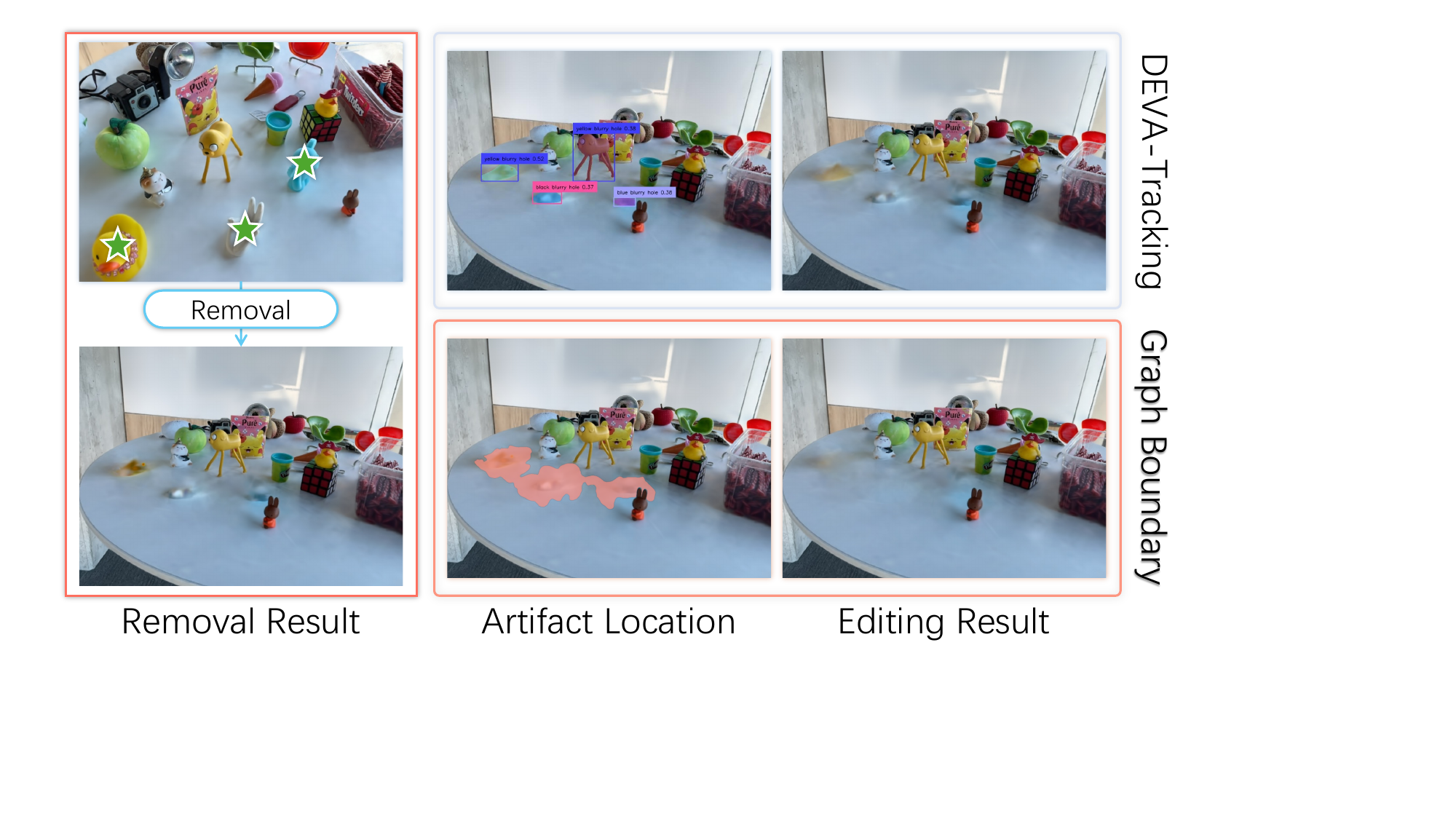}
  \caption{Object removing editing results with artifact regions localized and inpainted with different techniques.
  Compared to the DEVA-Tracking~\cite{cheng2023tracking} adopted by GsGrouping~\cite{gaussian_grouping} (top row), our anchor-graph structured representation (bottom row) enables an accurate localization of the artifact regions and thus makes realistic inpainting results without affecting the surrounding objects.
  }
  \label{fig:appendix_inpainting}
\end{figure}

\begin{figure}[!t]
  \centering
  \includegraphics[width=\linewidth]{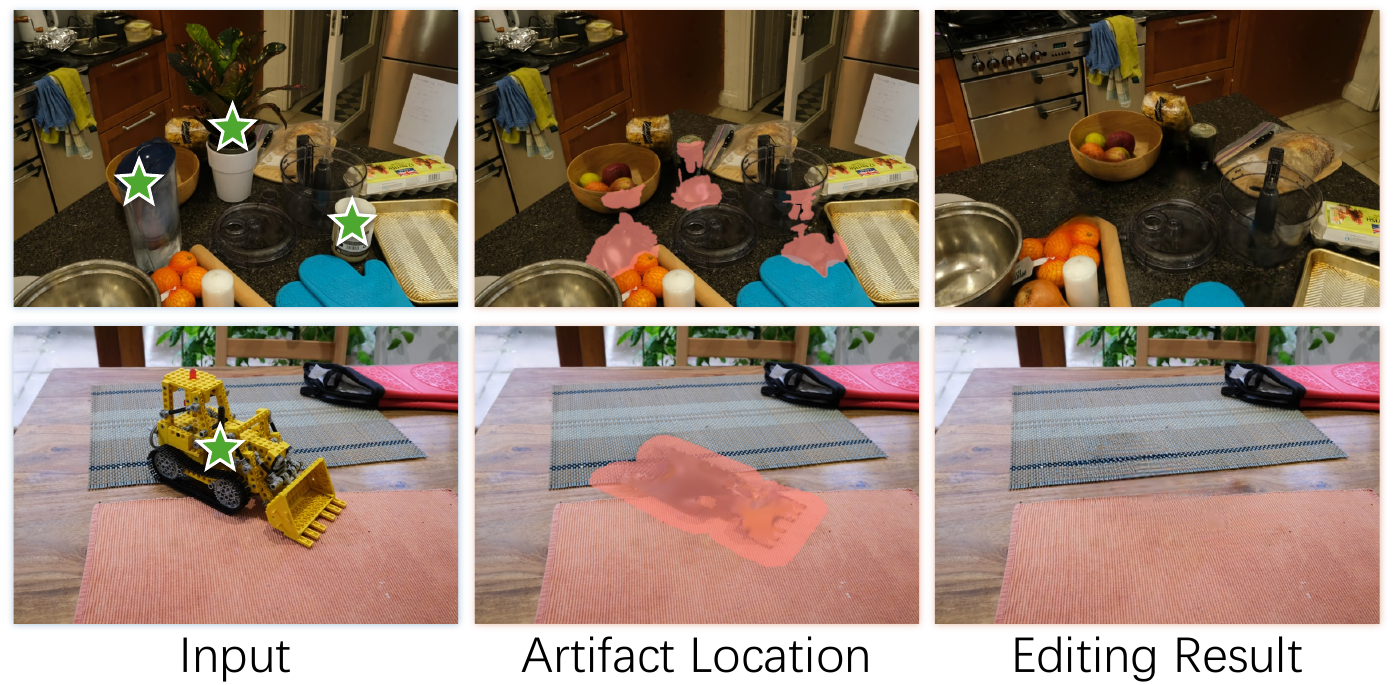}
  \caption{More editing results on MipNeRF360~\cite{barron2022mip} using our graph-based artifact localization technique.}
  \label{fig:appendix_more_inpainting}
\end{figure}

\begin{figure}[!t]
  \centering
  \includegraphics[width=0.7\linewidth]{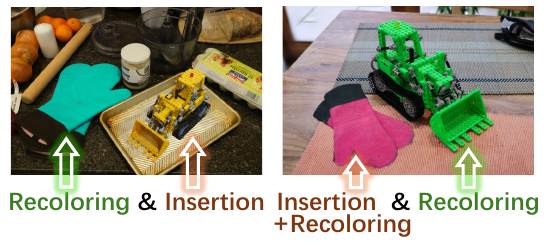}
  \vspace{-10pt}
  \caption{More object recoloring and insertion editing results on MipNeRF360~\cite{barron2022mip}.}
  \label{fig:appendix_more_insert_recolor}
  \vspace{-10pt}
\end{figure}

Directly removing the selected Gaussians for the objects makes artifacts in the remaining scene, due to the missing observations of the occluded region across all views, as shown in the left column of Figure~\ref{fig:appendix_inpainting}. Thus an inpainting operation is necessary to fill the holes.

We compare the two inpainting techniques adopted by GsGrouping~\cite{gaussian_grouping} and our approach, which differ in localizing the artifact regions to be repaired. GsGrouping uses Deva Tracking~\cite{cheng2023tracking}. As shown in the top row of the figure, due to ambiguous features and the difficulty of precisely identifying the hole regions, most viewpoints fail to maintain a stable artifact mask, resulting in suboptimal editing outcomes. By contrast, our anchor-graph structure enables an accurate selection of the object including the inner Gaussians, thus providing a precise localization of the artifact region by extending the boundary of the selected object, yielding more reliable and visually coherent editing results.


To further validate the performance of our artifact localization, we performed editing experiments on two scenes from the Mip-NeRF360 dataset~\cite{barron2022mip}. For the counter scene, we removed three objects of varying sizes, including a transparent kettle. As for the kitchen scene, we evaluated our method's ability to repair large hole regions resulting from object removal. As shown in Figure~\ref{fig:appendix_more_inpainting}, our approach accurately identifies and fills the hole regions, resulting in high-quality and consistent scene editing.

Additionally, we present the results of object recoloring and insertion of a complex scene in Figure~\ref{fig:appendix_more_insert_recolor}, which contains many objects close to each other and has occlusions across multiple views.

\section{Computation Overhead}\label{appendix:computation}

The maximum reserved memory, training time, and rendering FPS are reported in Table~\ref{tab:time_and_memory}.
For our anchor-graph structure, we store the anchors only for the occupied voxels and the sparse edges between neighbor anchors, incurring minimal additional memory.
On the other hand, this structure regularizes the Gaussian primitives to lie around the object surfaces, which largely reduces the number of Gaussians and thus the training time.
Notably, we do not intend to claim a faster rendering speed, since we implemented a CUDA-based module to render RGB, feature map, and other outputs in one pass, while SAGA and OpenGaussian need to invoke the renderer multiple times.
\begin{table}[!ht]
\small
\centering
\caption{Computation Overhead on LERF dataset~\cite{kerr2023lerf}.}
\begin{tabular}{c|ccc}
\toprule
Methods          & Memory$\downarrow$    & Train Time$\downarrow$  & Rendering FPS$\uparrow$ \\
\midrule
SAGA            & 13.29 GB              & \textbf{33.63 mins}        & $\sim$252     \\
GsGrouping      & 20.21 GB              & 51.21 mins        & $\sim$114     \\
OpenGaussian    & 16.81 GB              & 74.31 mins        & $\sim$96      \\
$w/\ codebook$  & 12.91 GB              & 69.97 mins        & $\sim$185     \\
Ours            & \textbf{7.56 GB}               & 39.55 mins        & \textbf{$\sim$515}     \\
\bottomrule
\end{tabular}
\label{tab:time_and_memory}
\end{table}

\begin{table*}[!ht]
\centering
\caption{Quantitative evaluation of text querying on Mip-NeRF360 dataset~\cite{barron2022mip}.
}
\resizebox{\linewidth}{!}{
\begin{tabular}{l|ccccccc|ccccccc}
\toprule
                         & \multicolumn{7}{c|}{mIoU. $\uparrow$}                                                    & \multicolumn{7}{c}{mBIoU. $\uparrow$}          \\
\multirow{-2}{*}{Methods} & \texttt{bicycle} & \texttt{bonsai} & \texttt{counter} & \texttt{garden} & \texttt{kitchen} & \texttt{room}  & \cellcolor[HTML]{EFEFEF}\textbf{Mean}  & \texttt{bicycle} & \texttt{bonsai} & \texttt{counter} & \texttt{garden} & \texttt{kitchen} & \texttt{room}  & \cellcolor[HTML]{EFEFEF}\textbf{Mean}  \\
\midrule
SAGA                     & 1.58    & 32.38  & 19.24   & 19.21  & 17.26   & 0.16  & \cellcolor[HTML]{EFEFEF}{14.97} & 2.13    & 24.21  & 15.68   & 15.36  & 9.33    & 0.2   & \cellcolor[HTML]{EFEFEF}11.15 \\
GsGrouping               & 10.52   & \textbf{68.73}  & 47.73   & \textbf{34.59}  & 61.7    & 41.22 & \cellcolor[HTML]{EFEFEF}{44.08} & 8.89    & \textbf{53.62}  & 44.61   & 29.52  & \textbf{54.52}   & 36.04 & \cellcolor[HTML]{EFEFEF}37.86 \\
OpenGaussian             & 25.97   & 33.2   & 47.52   & 25.87  & 41.42   & 41.7  & \cellcolor[HTML]{EFEFEF}{35.94} & 15.41   & 26.34  & 41.59   & 20.87  & 21.7    & 35.84 & \cellcolor[HTML]{EFEFEF}26.95 \\
Ours                     & \textbf{31.15}   & 53.47  & \textbf{61.89}   & 34.46  & \textbf{62.26}   & \textbf{50.76} & \cellcolor[HTML]{EFEFEF}{\textbf{48.99}} & \textbf{18.94}   & 48.49  & \textbf{58.85}   & \textbf{31.52}  & 41.77   & \textbf{45.03} & \cellcolor[HTML]{EFEFEF}\textbf{40.76} \\
\bottomrule
\end{tabular}
}
\label{tab:text_mip360}
\end{table*}
\begin{table*}[!ht]
\centering
\caption{Quantitative evaluation of click querying on LLFF dataset~\cite{mildenhall2019llff}.
}
\resizebox{\linewidth}{!}{
\begin{tabular}{l|ccccccccc|ccccccccc}
\toprule
                          & \multicolumn{9}{c|}{mIoU. $\uparrow$}                                                                                                                                                       & \multicolumn{9}{c}{mBIoU. $\uparrow$}                                                                                                                                                       \\
\multirow{-2}{*}{Methods} & \texttt{fern}           & \texttt{flower}         & \texttt{fortress}       & \texttt{horns\_c}  & \texttt{horns\_l}    & \texttt{leaves}         & \texttt{orchids}        & \texttt{trex}           & \cellcolor[HTML]{EFEFEF}Mean           & \texttt{fern}           & \texttt{flower}         & \texttt{fortress}       & \texttt{horns\_c}  & \texttt{horns\_l}    & \texttt{leaves}         & \texttt{orchids}        & \texttt{trex}           & \cellcolor[HTML]{EFEFEF}Mean           \\
\midrule
SAGA                      & \textbf{82.53} & 95.15          & 98.15          & 92.83          & 94.57          & 92.88          & 88.82          & 83.99          & \cellcolor[HTML]{EFEFEF}91.61          & 75.12          & 80.87          & 78.18          & 68.44          & 72.2           & 77.89          & 74.76          & 70.25          & \cellcolor[HTML]{EFEFEF}75.04          \\
GsGrouping                & 80.70          & 57.72          & 97.75          & 96.78          & 94.58          & 70.5           & 36.13          & 51.69          & \cellcolor[HTML]{EFEFEF}72.73          & 64.74          & 35.99          & 55.02          & 69.74          & 73.57          & 48.68          & 26.09          & 49.38          & \cellcolor[HTML]{EFEFEF}52.56          \\
OpenGaussian              & 70.74          & 62.63          & 94.91          & 79.81          & 77.81          & 87.68          & 59.88          & 68.88          & \cellcolor[HTML]{EFEFEF}75.29          & 58.81          & 36.75          & 67.38          & 47.24          & 52.85          & 57.81          & 43.49          & 66.25          & \cellcolor[HTML]{EFEFEF}53.82          \\
Ours                      & 82.01          & \textbf{95.38} & \textbf{98.59} & \textbf{97.36} & \textbf{96.31} & \textbf{93.89} & \textbf{90.76} & \textbf{87.02} & \cellcolor[HTML]{EFEFEF}\textbf{92.66} & \textbf{77.85} & \textbf{81.73} & \textbf{91.06} & \textbf{81.24} & \textbf{83.54} & \textbf{80.71} & \textbf{80.42} & \textbf{85.24} & \cellcolor[HTML]{EFEFEF}\textbf{82.64}\\
\bottomrule
\end{tabular}
}
\label{tab:click_llff}
\end{table*}

\section{More Ablation Study Results}\label{appendix:ab}

\begin{figure*}[!ht]
  \centering
  \includegraphics[width=\linewidth]{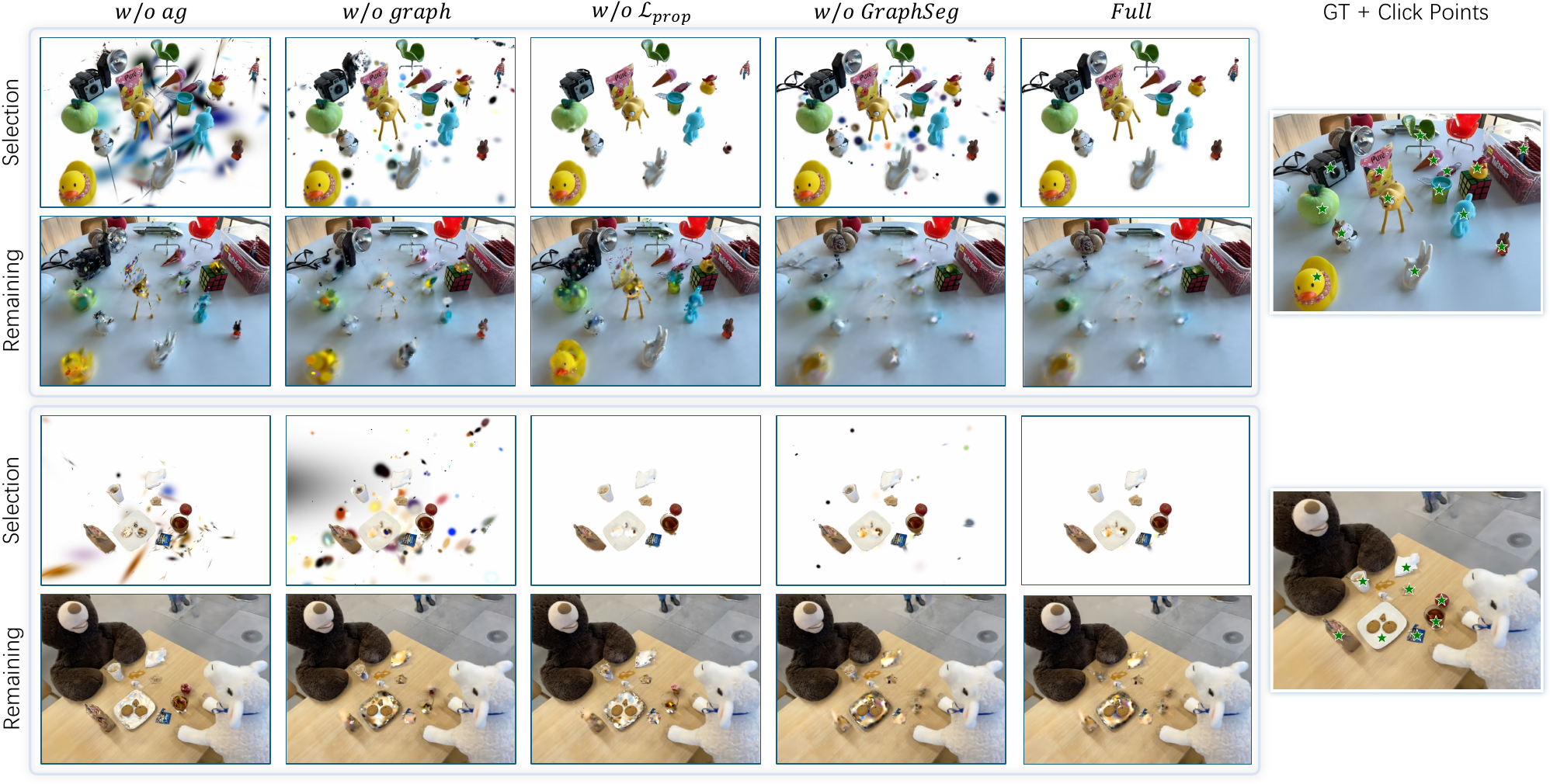}
  \caption{Ablation study results. We separately validated the importance of our key design for segmentation task, the Anchor-Gaussian structure, and the Graph-based operation. The advantage of the Anchor-Gaussian is demonstrated by comparing $w/o\ GraphSeg$ with a variant that uses 3DGS without our anchor-graph ($w/o\ ag$). The effectiveness of our Graph-based Operation respectively adopting $w/o\ graph$, $w/o\ \mathcal{L}_{prop}$ and $w/o\ GraphSeg$.}
  \label{fig:appendix_ab}
\end{figure*}

\begin{table*}[ht]
\centering
\caption{Full ablation studies on the LERF-OVS dataset~\cite{kerr2023lerf} about the key designed.
}
\resizebox{\linewidth}{!}{
\begin{tabular}{c|cc|ccccc|ccccc}
\toprule
&&& \multicolumn{5}{c|}{mIoU $\uparrow$} 
& \multicolumn{5}{c}{mBIoU. $\uparrow$} \\
\multirow{-2}{*}{Case} & $w/\ \mathcal{L}_{prop}$ & $w/\ GraphSeg$ &
  \texttt{figurines} &
  \texttt{teatime} &
  \texttt{ramen} &
  \texttt{kitchen} &
  \cellcolor[HTML]{EFEFEF}\textbf{Mean} &
  \texttt{figurines} &
  \texttt{teatime} &
  \texttt{ramen} &
  \texttt{kitchen} &
  \cellcolor[HTML]{EFEFEF}\textbf{Mean} \\ 
  \midrule
\#1  &                          &                   
& 57.62& 64.72&  26.39& 22.14& \cellcolor[HTML]{EFEFEF} 42.72&  56.72& 61.30& 26.05&  16.73& \cellcolor[HTML]{EFEFEF} 40.20\\
\#2  &                          & \Checkmark        
& 55.95& 66.54&  31.45& 29.50& \cellcolor[HTML]{EFEFEF} 45.85&  58.59& 63.13& 31.02&  21.32& \cellcolor[HTML]{EFEFEF} 43.51 \\
\#3  & \Checkmark               &    
& 65.08& 71.16&  28.15& \textbf{32.01}& \cellcolor[HTML]{EFEFEF} 49.10&  63.61& 67.33& 26.46&  21.46& \cellcolor[HTML]{EFEFEF} 44.72\\
Full & \Checkmark               & \Checkmark        &

  \textbf{66.98} &
  \textbf{71.62} &
  \textbf{47.99} &
  {30.82} &
  \cellcolor[HTML]{EFEFEF}\textbf{54.35} &
  \textbf{65.30} &
  \textbf{67.83} &
  \textbf{42.45} &
  \textbf{22.15} &
  \cellcolor[HTML]{EFEFEF}\textbf{49.43} \\
  \bottomrule
\end{tabular}
}
\label{tab:appendix_lerf}
\end{table*}

Table~\ref{tab:appendix_lerf} presents the complete ablation study results on the LERF dataset~\cite{kerr2023lerf}. Overall, our graph-related operations significantly improve both mask completeness and boundary quality, as evidenced by notable gains in mIoU and mBIoU.

To further assess the importance of these operations for the query task, we demonstrate the selected Gaussians and the remaining scenes.
Figure~\ref{fig:appendix_ab} provides a full visualization of all ablation variants.
Our graph-based region growing effectively prevents the selection of Gaussians outside the target object, as demonstrated by the comparison between the $w/o\ GraphSeg$ variant and our full method.
Moreover, our graph propagation smooths the feature field within the object and enhances a clean Gaussian selection, effectively eliminating inner Gaussians in the remaining scenes, as shown by the comparison between $w/o\ \mathcal{L}_{prop}$ and our full method.
Additionally, our anchor-Gaussian structure effectively constrains the local distribution of Gaussians, as demonstrated by the comparison between $w/o\ ag$ and $w/o\ Graph$.
Overall, our full method not only enables the clean selection of objects but also ensures the comprehensive inclusiveness of the inner object Gaussians.

\section{Comparison with Other Structured‑GSes}\label{appendix:structured}

Several recent works explore structured 3DGS, but for different goals and thus framework designs.
Scaffold‑GS~\cite{scaffoldgs} proposes the Anchor-Gaussian structure to distribute local 3D Gaussians and predicts their view-adaptive attributes. However, it does not localize the Gaussians to distribute within the voxel of the corresponding anchor, and eliminates the anchor-graph for the feature propagation. SuperGSeg~\cite{liang2024supergsegopenvocabulary3dsegmentation} proposes to cluster the optimized Gaussians into Super-Gaussians and distill the semantic features to comprehensively understand 3D scenes. However, it lacks anchor-graph-based propagation to further refine the local feature fields and requires a much larger memory cost during training. 

Therefore, we perform the ablation study experiments ($w/o\ localization$ and $w/\ codebook$) to validate the effectiveness of our design, as shown in Table~\ref{tab:more_ab}. Specifically, for $w/o\ localization$, we remove the scaling constraint (Eq.~2)  and structured spatial regularization (Eq.~3-4), to evaluate the effectiveness of our anchor-graph structure compared to ScaffoldGS and SuperGSeg. For $w/\ codebook$, we preserve our stage 1 and introduce a learnable codebook to emulate the Super-Gaussians proposed by SuperGSeg. Our full approach significantly outperforms both variants in segmentation accuracy, demonstrating the advantages of our anchor-graph–based localization and propagation.

\begin{table*}[!ht]
\small
\centering
\caption{Ablation Study of Structured-GS Design on LERF-OVS dataset~\cite{kerr2023lerf}}

\resizebox{\linewidth}{!}{
\begin{tabular}{c|ccccc|ccccc}
\toprule
& \multicolumn{5}{c|}{mIoU $\uparrow$} 
& \multicolumn{5}{c}{mBIoU. $\uparrow$} \\
\multirow{-2}{*}{Methods}&
  \texttt{figurines} &
  \texttt{teatime} &
  \texttt{ramen} &
  \texttt{kitchen} &
  \cellcolor[HTML]{EFEFEF}\textbf{Mean} &
  \texttt{figurines} &
  \texttt{teatime} &
  \texttt{ramen} &
  \texttt{kitchen} &
  \cellcolor[HTML]{EFEFEF}\textbf{Mean} \\ 
  \midrule

$w/\ codebook$    & 49.83 & 66.77 &  17.99 & 27.48 & \cellcolor[HTML]{EFEFEF} 40.51 & 35.33 & 60.67 & 15.29 & 20.3 & \cellcolor[HTML]{EFEFEF} 32.89 \\
$w/o\ localized$  & 28.40 & 53.94 &  14.51 & 24.97 & \cellcolor[HTML]{EFEFEF} 30.45 & 25.94 & 51.39 & 13.04 & 19.73 & \cellcolor[HTML]{EFEFEF} 27.52 \\
Ours &
  \textbf{66.98} &
  \textbf{71.62} &
  \textbf{47.99} &
  \textbf{30.82} &
  \cellcolor[HTML]{EFEFEF}\textbf{54.35} &
  \textbf{65.30} &
  \textbf{67.83} &
  \textbf{42.45} &
  \textbf{22.15} &
  \cellcolor[HTML]{EFEFEF}\textbf{49.43} \\
  \bottomrule
\end{tabular}
}
\label{tab:more_ab}
\end{table*}

\end{document}